\documentclass[lettersize,journal]{IEEEtran}
\usepackage{amsmath,amsfonts}
\usepackage{algorithmic}
\usepackage{algorithm}
\usepackage{caption}
\usepackage{subcaption}
\usepackage{array}
\usepackage[caption=false,font=normalsize,labelfont=sf,textfont=sf]{subfig}
\usepackage{textcomp}
\usepackage{stfloats}
\usepackage{url}
\usepackage{verbatim}
\usepackage{graphicx}
\usepackage{cite}
\usepackage{color}
\usepackage{multirow}
\usepackage{bbm}
\usepackage{colortbl}
\usepackage{tcolorbox}
\usepackage{bbding}
\usepackage{titlesec}
\usepackage{wrapfig}
\usepackage{threeparttable}
\usepackage{booktabs}
\usepackage{makecell}
\usepackage{adjustbox}
\usepackage[switch]{lineno}
\begin{document}

\title{Synergistic Dual-Branch Adaptation for Multi-modal Generalized Category Discovery}

\author{Yuxun~Qu$^{\dagger}$,  
        Minyu~Zhou$^{\dagger}$,
        Yongqiang~Tang,
        Chenyang~Zhang
        and~Wensheng~Zhang
\thanks{$^{\dagger}$ These authors contributed equally to this work.}
\thanks{ Yuxun Qu and Minyu Zhou are with the College of Computer Science, Nankai University, Tianjin 300350, China. (e-mail: quyuxun97@gmail.com, minyu.zhou@mail.nankai.edu.cn)}
\thanks{ Yongqiang Tang and Chenyang Zhang are with the State Key Laboratory of Multimodal Artificial Intelligence Systems, Institute of Automation, Chinese Academy of Sciences, Beijing 101408, China. (e-mail: yongqiang.tang@ia.ac.cn, chenyang.zhang@ia.ac.cn)}
\thanks{ Wensheng Zhang is with the College of Computer Science, Nankai University, Tianjin 300350, China, and also with the State Key Laboratory of Multimodal Artificial Intelligence Systems, Institute of Automation, Chinese Academy of Sciences, Beijing 101408, China. (e-mail: zhangwenshengia@hotmail.com).}}

\markboth{Journal of \LaTeX\ Class Files,~Vol.~14, No.~8, August~2021}%
{Shell \MakeLowercase{\textit{et al.}}: A Sample Article Using IEEEtran.cls for IEEE Journals}

\IEEEpubid{0000--0000/00\$00.00~\copyright~2021 IEEE}

\maketitle

\begin{abstract}
Generalized Category Discovery (GCD) aims to classify old categories and discover new ones from unlabeled data. Recent multi-modal approaches introduce retrieved or synthesized texts into a dual-branch architecture to provide semantic cues complementary to visual features. However, the cross-modal synergy in existing dual-branch methods remains coarse and incomplete: the two modalities are encoded independently with the bias and noise in the derived text left unaddressed during encoding, and existing mutual learning strategies operate only on global class-level anchors, lacking fine-grained relational supervision. To address these limitations, we propose the Synergistic Dual-Branch Adaptation (SDBA) framework, which serves as a plug-and-play enhancement compatible with existing dual-branch methods such as GET and TextGCD. SDBA comprises two components: the cross-modal synergistic adapter inserts lightweight adapters into both branches and further injects visual information into the text adapter at each encoder layer to enhance text feature learning during encoding; the neighborhood mutual learning module enforces consistent local neighborhood distributions between the two branches via bidirectional KL divergence, providing fine-grained relational supervision for both old and new classes. Extensive experiments on six benchmarks demonstrate state-of-the-art performance, and consistent improvements on different baselines validate the broad scalability of the proposed framework.
\end{abstract}


\section{Introduction}
\label{sec:intro}

In real-world image classification, Semi-Supervised Learning (SSL)~\cite{DBLP:journals/tcsv/JiaLLH23, DBLP:conf/nips/LiWLYY0L23} effectively reduces annotation cost by exploiting unlabeled images. However, SSL assumes a closed-world setting where the unlabeled data shares the same label space as the labeled set~\cite{DBLP:conf/iclr/LaineA17}, limiting its applicability when new categories continuously emerge. 
To relax this constraint, Generalized Category Discovery (GCD)~\cite{DBLP:conf/cvpr/gcdVazeHVZ22} assumes that the unlabeled dataset contains new categories absent from the labeled set. The goal of GCD is not only to classify old categories but also to discover and cluster new categories in the unlabeled data. As a promising direction with broad practical relevance, GCD has attracted growing research interest.

\IEEEpubidadjcol
 
\begin{figure}[t]
  \centering
   \includegraphics[width=1\linewidth]{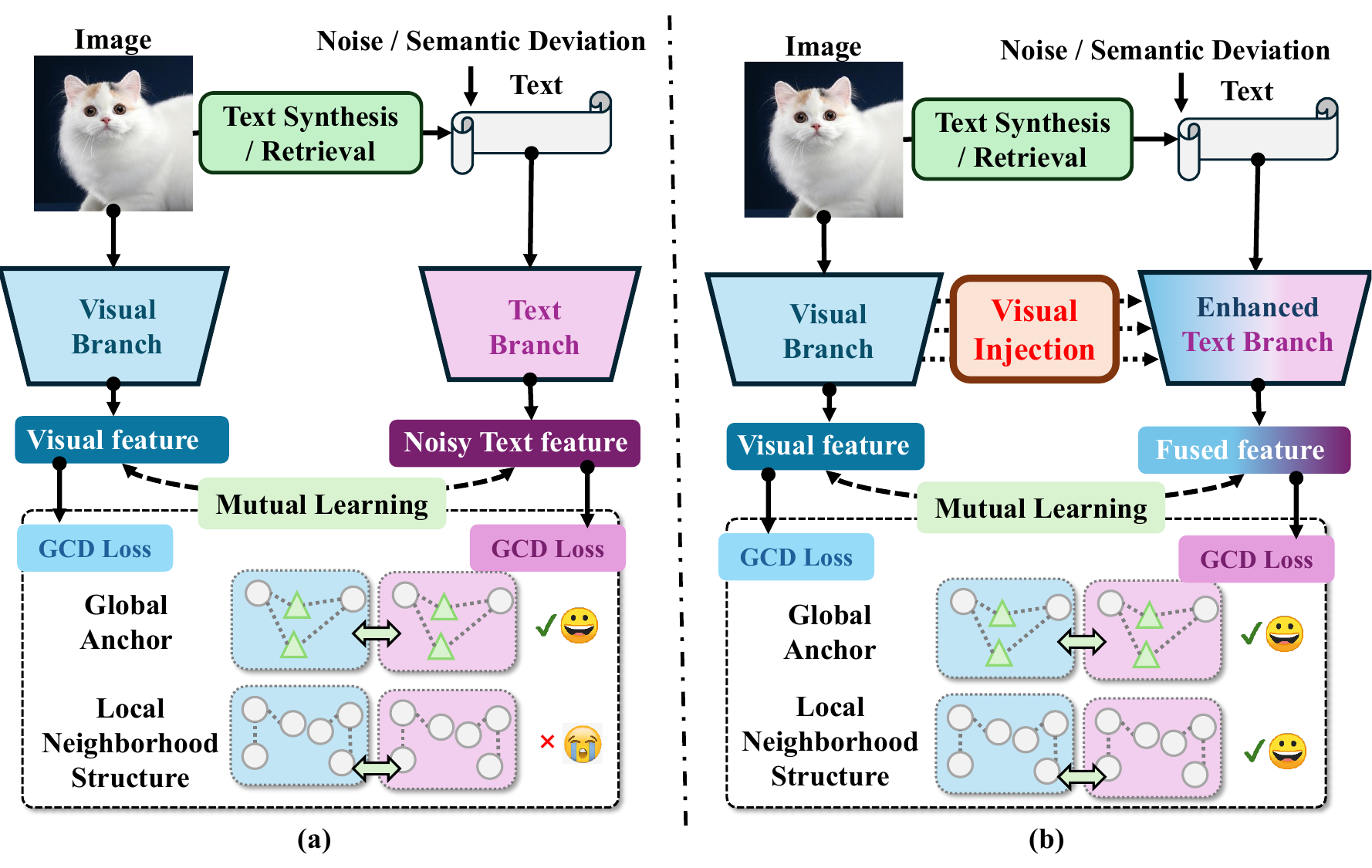}
   \caption{
   \textbf{Motivation.} (a) Conventional dual-branch methods encode the two modalities independently and rely on global class anchors in mutual learning, leaving the bias and noise in the retrieved or synthesized texts propagating through the cross-modal loss and degrading the learning of both branches.  (b) The proposed SDBA injects visual class representations into the text adapter and enables neighborhood mutual learning, enhancing the text branch and promoting mutual reinforcement.}
   \label{fig:motivation}
\end{figure}

While traditional GCD methods rely solely on visual features~\cite{DBLP:conf/iccv/simgcdWenZQ23, DBLP:conf/cvpr/gcdVazeHVZ22}, the text modality provides discriminative cues that are complementary to visual appearance, motivating the incorporation of multi-modal information into GCD. {\color{black}Recent dual-branch approaches like TextGCD~\cite{DBLP:conf/eccv/ZhengPLSZ24} and GET~\cite{DBLP:conf/cvpr/WangPXYLC25}} introduce multi-modal pre-trained models (e.g., CLIP) into the GCD framework. These methods acquire auxiliary text representations via retrieval~\cite{DBLP:journals/corr/abs-2305-10420, DBLP:conf/eccv/ZhengPLSZ24} or synthesis~\cite{DBLP:conf/cvpr/WangPXYLC25} in the absence of paired text annotations, 
and perform intra-modal category discovery within each branch, with cross-modal mutual learning enabling synergy between the two modalities. By virtue of their simplicity and effectiveness, dual-branch methods have become the dominant paradigm for multi-modal GCD.

However, as illustrated in Fig.~\ref{fig:motivation}, current dual-branch approaches achieve only coarse and incomplete cross-modal synergy throughout the learning pipeline, hindering the full exploitation of cross-modal information in GCD and manifesting as two interrelated limitations.
\textbf{(1) Isolated Dual-Branch Encoding.}  
In GCD, since paired image-text supervision is unavailable, text information is often retrieved from a limited vocabulary or synthesized via an auxiliary generator~\cite{DBLP:conf/eccv/ZhengPLSZ24, DBLP:conf/cvpr/WangPXYLC25}, inevitably introducing bias and noise into the text branch. 
However, existing methods encode the two modalities separately and only interact through a mutual learning loss, which fails to mitigate such noise during encoding. The noisy supervision signal from the text branch then propagates to the visual branch through the cross-modal loss, degrading the learning of both branches.
\textbf{(2) Insufficient Structural Mutual Learning.} Current mutual learning strategies enforce cross-modal consistency at the level of fixed global anchors such as class prototypes or class centers. Each sample is represented by its similarity to these anchors, and the resulting similarity distributions are aligned across modalities. Such anchor-based alignment captures only coarse class-level relations and fails to model the fine-grained local neighborhood structure among samples. 
This limitation is particularly severe for new-class samples, as their similarity to any fixed anchor tends to be low, rendering the mutual learning signal uninformative. 
{\color{black}This motivates us to introduce finer-grained constraints to complement the cross-modal interaction.}

To address these limitations, we propose the \textbf{\underline{S}}ynergistic \textbf{\underline{D}}ual-\textbf{\underline{B}}ranch \textbf{\underline{A}}daptation \textbf{(SDBA)} framework, which replaces the conventional independent dual-branch structure with a synergistically coupled architecture comprising two core components: the \textbf{\underline{C}}ross-\textbf{\underline{M}}odal \textbf{\underline{S}}ynergistic \textbf{\underline{A}}dapter \textbf{(CMSA)} and \textbf{\underline{N}}eighborhood \textbf{\underline{M}}utual \textbf{\underline{L}}earning \textbf{(NML)}. To address the first limitation, CMSA adopts lightweight adapters with frozen backbones for parameter-efficient adaptation, and further incorporates a visual injection mechanism into the text adapter to mitigate the bias and noise in the derived text. Specifically, the global class representation of the visual branch is fused with the text feature at each encoder layer, enabling layer-wise enhancement of text features with visual context. In addition, NML replaces fixed global anchors with local sample neighborhoods as the basis for cross-modal alignment. For each sample, the neighborhood is constructed from its similar samples in feature space of each modality, and the resulting neighborhood distributions are aligned across branches via bidirectional KL divergence, providing fine-grained relational supervision for both old and new classes.
 
In summary, our contributions are listed as follows.
\begin{itemize}
    \item {\color{black}
    We propose the Synergistic Dual-Branch Adaptation (SDBA) framework, which mitigates the coarse and insufficient cross-modal synergy in both encoding and mutual learning through the cross-modal synergistic adapter and the neighborhood mutual learning module.}
    \item We propose the cross-modal synergistic adapter, which injects visual information into the adapter in the text branch to suppress the noise introduced during text acquisition. 
    \item We introduce neighborhood mutual learning, which utilizes local neighborhood distributions instead of relying on fixed global anchors for cross-modal mutual learning, thereby enabling fine-grained interaction. 
    \item Extensive experiments on six benchmarks demonstrate that our method achieves state-of-the-art performance. Furthermore, the consistent performance gains achieved when integrating our components into 
    the existing dual-branch methods validate the broad scalability of the proposed approach.
\end{itemize}

The rest of the paper is organized as follows. Section~\ref{section:related work} briefly reviews related work. Section~\ref{sec:preliminary} introduces the preliminaries. Section~\ref{section:proposal} elaborates on the proposed method. Extensive experiments and analyses are presented in Section~\ref{section:experiment}. Finally, Section~\ref{section:conclusion} concludes the paper.

\section{Related Work}
\label{section:related work}
 
\subsection{Semi-Supervised Learning}

Semi-supervised learning (SSL)~\cite{DBLP:journals/tcsv/ChenWWPIL18, DBLP:journals/tcsv/TangLNZ23,DBLP:journals/tmm/HuCCZG25,DBLP:journals/tmm/WangWWHL24,DBLP:journals/tmm/ZhangZWYWWLG23} alleviates the reliance on costly annotation by exploiting unlabeled data jointly with a small labeled set, and is broadly categorized into two paradigms: consistency regularization and pseudo-labeling. Consistency regularization~\cite{DBLP:conf/iclr/LaineA17, DBLP:conf/nips/TarvainenV17} requires stable predictions under different perturbations or training stages, with representative works such as Temporal Ensembling~\cite{DBLP:conf/iclr/LaineA17} and Mean Teacher~\cite{DBLP:conf/nips/TarvainenV17} producing smoother pseudo-supervision via exponential moving average. Pseudo-labeling strategies~\cite{DBLP:conf/ijcnn/ArazoOAOM20, DBLP:conf/iclr/RizveDRS21} instead assign hard or soft labels to unlabeled samples and optimize them jointly with the supervised objective. FixMatch~\cite{DBLP:conf/nips/SohnBCZZRCKL20} unifies the two paradigms by enforcing consistency between weakly and strongly augmented views, while follow-ups such as MixMatch~\cite{DBLP:conf/nips/BerthelotCGPOR19} and FlexMatch~\cite{DBLP:conf/nips/ZhangWHWWOS21} further improve performance via mixed interpolation and adaptive thresholding. Despite these advances, mainstream SSL methods still operate under the closed-world assumption, limiting their applicability in open-world settings.

\subsection{Generalized Category Discovery}
 
Generalized Category Discovery~\cite{DBLP:conf/cvpr/gcdVazeHVZ22,DBLP:journals/tmlr/FathalizadehR26,peng2026sharpness,wang2026learning} extends SSL to an open-world setting where the unlabeled set contains both old and new categories. Early methods~\cite{DBLP:conf/cvpr/gcdVazeHVZ22} rely on contrastive learning with a non-parametric classifier, establishing the basic learning paradigm. SimGCD~\cite{DBLP:conf/iccv/simgcdWenZQ23} introduces a parametric classifier with self-distillation, substantially boosting performance and becoming the dominant baseline. Subsequent works advance along two directions. On the learning objective side, $\mu$GCD~\cite{DBLP:conf/nips/VazeVZ23} adopts Mean Teacher to produce more stable training targets, and LegoGCD~\cite{DBLP:conf/cvpr/CaoZWYS0L024} identifies catastrophic forgetting of old classes as a key failure mode of SimGCD and introduces local entropy regularization to preserve old-class knowledge during new-class learning. On the model adaptation side, PromptCAL~\cite{DBLP:conf/cvpr/promptcalgcdZhangKSNCK23} and SPTNet~\cite{wang2024sptnet} develop prompt-based strategies to better elicit the pretrained visual knowledge of the backbone. AdaptGCD~\cite{11142371} is the first to introduce adapter tuning into GCD, freezing the backbone and inserting lightweight bottleneck modules in parallel with the feed-forward layers for parameter-efficient adaptation, demonstrating a favorable balance between generalization capacity and task adaptability. Despite these advances, all the above methods rely solely on visual features and are limited in their ability to distinguish categories with high inter-class visual similarity.
 
\subsection{Multi-Modal Generalized Category Discovery}

Since unimodal GCD methods are limited in distinguishing visually similar categories, a line of multi-modal GCD methods incorporates the text modality to provide complementary semantic cues~\cite{DBLP:conf/cvpr/SuZH0WW025}. As class names for unlabeled data are unavailable, two strategies are commonly adopted to obtain text: retrieval-based approaches that retrieve descriptive texts from a pre-built corpus, and synthesis-based approaches that generate text embeddings directly from visual features. ClipGCD~\cite{DBLP:journals/corr/abs-2305-10420} first introduces text into GCD by retrieving top-$k$ relevant texts from an external corpus via CLIP-based image-text similarity, and concatenating the retrieved text embeddings with visual features for joint semi-supervised clustering. Built upon this retrieval line, TextGCD~\cite{DBLP:conf/eccv/ZhengPLSZ24} constructs an LLM-enhanced visual lexicon and adopts a dual-branch framework with cross-modal co-teaching to better leverage the retrieved texts. In parallel, GET~\cite{DBLP:conf/cvpr/WangPXYLC25} explores the synthesis route by introducing a text embedding synthesizer that converts visual features into synthetic text embeddings, removing the dependency on external corpora. However, these dual-branch methods encode the two modalities independently and interact only through a cross-modal loss against fixed global anchors, failing to mitigate the noise in the text branch during encoding and providing only coarse-grained cross-modal supervision, which motivates this work.

\subsection{Visual-Language Models (VLMs)}

Visual-Language Models aim to align images and text in a shared embedding space, enabling cross-modal understanding and zero-shot generalization. CLIP~\cite{DBLP:conf/icml/RadfordKHRGASAM21} is the most representative work, which learns to align visual and textual representations via contrastive learning on large-scale web-crawled image-text pairs. Follow-up works such as SigLIP~\cite{zhai2023sigmoid} further improve training efficiency by replacing the softmax-based contrastive loss with a pairwise sigmoid loss. Building on the semantic alignment from CLIP, several methods~\cite{Menon2022Visual, Novack2023Chils, Yang2023Language} query language models to generate class-level textual descriptions as additional classification signals, achieving improvements in fine-grained and low-shot settings. However, all these methods presuppose the availability of class names at training time, leaving the use of VLMs for generalized category discovery, where new class names are entirely unavailable, an open challenge~\cite{DBLP:journals/ijcv/GaoGZCMZLS24, DBLP:conf/cvpr/YaoZX23, DBLP:journals/tpami/ZhangHJS24, DBLP:journals/tmm/WangXLLZJ23}.

\section{Preliminaries}
\label{sec:preliminary}
 
\subsection{Problem Formulation}

In Generalized Category Discovery (GCD), the dataset $\mathcal{D}$ consists of a labeled set $\mathcal{D}_l = \{(\mathbf{x}_i^{\rm v}, y_i)\}_{i=1}^{N}$ and an unlabeled set $\mathcal{D}_u = \{\mathbf{x}_i^{\rm v}\}_{i=1}^{M}$, where $\mathbf{x}_i^{\rm v} \in \mathcal{X}$ denotes the $i$-th image and $y_i \in \mathcal{Y}_l$ is its class label. The label space of $\mathcal{D}_l$ is $\mathcal{Y}_l$, referred to as \emph{known classes} (or \emph{old classes}). The label space of $\mathcal{D}_u$ is $\mathcal{Y}_u$, which satisfies $\mathcal{Y}_l \subset \mathcal{Y}_u$; the subset $\mathcal{Y}_u \setminus \mathcal{Y}_l$ constitutes the \emph{novel classes} (or \emph{new classes}) that are absent from $\mathcal{D}_l$. The goal is to classify samples in $\mathcal{D}_u$ belonging to $\mathcal{Y}_l$ and simultaneously discover and cluster those belonging to $\mathcal{Y}_u \setminus \mathcal{Y}_l$.

\subsection{Dual-Branch Framework}
\label{sec:dualbranchframe}

The dual-branch framework~\cite{ DBLP:conf/cvpr/WangPXYLC25, DBLP:conf/eccv/ZhengPLSZ24} is the dominant paradigm in multi-modal generalized category discovery and serves as the foundation of our method. By introducing the text modality as an auxiliary signal, it constructs a dual-stream architecture with three components: text information acquisition, intra-modal category discovery, and cross-modal mutual learning.

{\color{black}
\emph{1) Text Information Acquisition.}
Since ground-truth captions for unlabeled images are unavailable, auxiliary text representations are constructed from visual inputs as $\mathbf{x}_i^{\rm t} = h(\mathbf{x}_i^{\rm v})$, where $h(\cdot)$ denotes the modality conversion operator. Synthesis-based methods (e.g., GET) build a lightweight mapping network to project visual features into the token embedding space, producing continuous synthetic text embeddings, while retrieval-based methods (e.g., TextGCD) use cosine-similarity-based retrieval over a pre-built visual lexicon to form discrete text inputs. Both require no ground-truth labels for unlabeled data.

\emph{2) Intra-Modal Category Discovery.}
Both branches perform intra-modal category discovery following SimGCD~\cite{DBLP:conf/iccv/simgcdWenZQ23}. For each branch, the loss combines a representation learning term $\mathcal{L}_{\rm rep}$ that integrates self-supervised and supervised contrastive learning, and a classification term $\mathcal{L}_{\rm cls}$ that is implemented as supervised loss and self-distillation loss with mean-entropy regularization. The per-branch loss is denoted as $\mathcal{L}_{\rm db}^{\rm v} = \mathcal{L}_{\rm rep}^{\rm v} + \mathcal{L}_{\rm cls}^{\rm v}$ for the visual branch and $\mathcal{L}_{\rm db}^{\rm t} = \mathcal{L}_{\rm rep}^{\rm t} + \mathcal{L}_{\rm cls}^{\rm t}$ for the text branch, with complete formulations provided in the Appendix. {\color{black}This objective achieves category discovery within each branch.}

\emph{3) Cross-Modal Mutual Learning.}
To bridge the two branches, a cross-modal mutual learning loss $\mathcal{L}_{\rm ml}$ is imposed. Taking CICO in GET~\cite{DBLP:conf/cvpr/WangPXYLC25} as an example, $\mathcal{L}_{\rm ml}$ computes per-sample similarity vectors $\mathbf{s}_i^{\rm v}, \mathbf{s}_i^{\rm t}$ between the sample features and the old-class prototype matrices $\mathbf{P}^{\rm v}, \mathbf{P}^{\rm t}$ in each modality, and aligns them via KL divergence (see Appendix for full formulation). TextGCD~\cite{DBLP:conf/eccv/ZhengPLSZ24} instead exchanges predictions between the two branches via co-teaching, with a multi-modal contrastive loss as an auxiliary objective. Both strategies rely on fixed global anchors (old-class prototypes or class centers), lacking fine-grained characterization of inter-sample relational structure.
 }
\section{Methodology}
\label{section:proposal}

\begin{figure*}[t]
  \centering
   \includegraphics[width=1\linewidth]{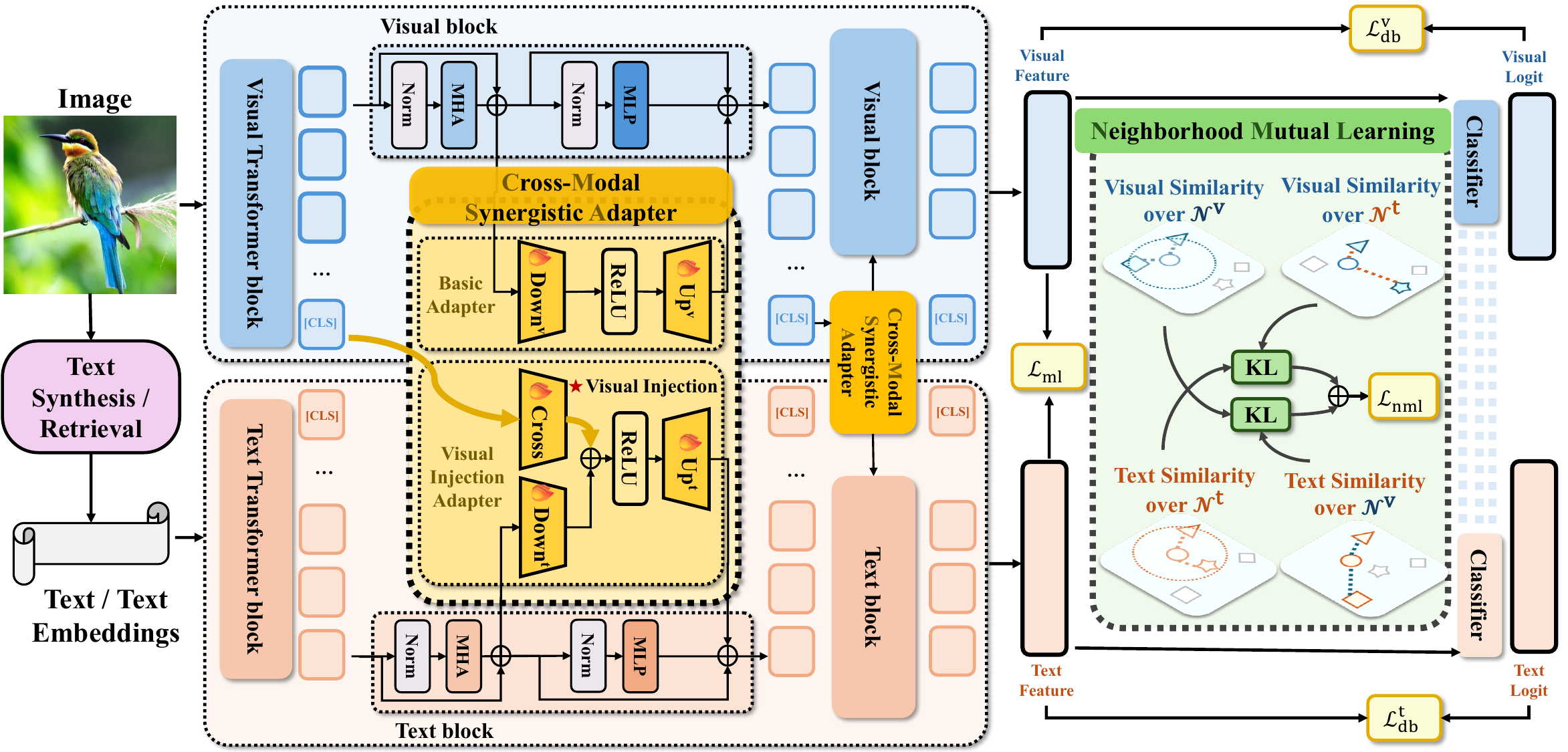}
   \caption{
   Framework overview of SDBA for multi-modal generalized category discovery. SDBA consists of two components: (i) a cross-modal synergistic adapter that inserts adapters into both branches and injects visual class representations into the text adapter in a layer-by-layer manner; and (ii) neighborhood mutual learning that enforces consistency between visual and text neighborhood distributions via bidirectional KL divergence. Here, $\mathcal{N}^{\rm v}$ and $\mathcal{N}^{\rm t}$ denote the visual and text neighborhood sets.
   }
   \label{fig:framework}
\end{figure*}

We propose the {synergistic dual-branch adaptation} framework, illustrated in Fig.~\ref{fig:framework}. Building upon the existing dual-branch paradigm, it introduces two novel components: the {cross-modal synergistic adapter} and the {neighborhood mutual learning} module. The {cross-modal synergistic adapter} achieves cross-modal synergy by injecting visual context into the text branch for layer-wise enhancement during encoding; the neighborhood mutual learning strengthens fine-grained cross-modal consistency by enforcing consistent neighborhood distributions between the two branches.

\subsection{Cross-Modal Synergistic Adapter (CMSA)}
\label{sec:cmsa}
 
{
The Cross-Modal Synergistic Adapter (CMSA) consists of two adapter modules tailored to the two modalities. For the visual branch, a basic adapter is inserted in each Transformer block. For the text branch, the basic adapter is enhanced with a visual injection mechanism that fuses the visual class representation into the low-dimensional bottleneck of the text adapter at each block. By injecting reliable visual context, the text branch is continuously guided during encoding, which mitigates the noise in the derived text and achieves deep cross-modal synergy. Notably, since the injection operates only from visual to text, the visual branch remains fully independent at inference.
}

\emph{1) Basic Adapter in the Visual Branch.} Following the architecture of AdaptFormer~\cite{chen2022adaptformer}, we introduce a lightweight bottleneck module in parallel with the frozen Feed-Forward Network (FFN) of each Transformer block. Eq.~(\ref{eq:vis_transf}) describes the forward process of FFN layer in the $l$-th visual Transformer block after inserting the adapter. 
\begin{equation}
    \mathbf{x}_{l+1,i}^{\rm v} = \mathrm{MLP}(\mathrm{LN}(\widetilde{\mathbf{x}}_{l,i}^{\rm v})) + \widetilde{\mathbf{x}}_{l,i}^{\rm v} + \Delta\mathbf{x}_{l,i}^{\rm v},
    \label{eq:vis_transf}
\end{equation}
where $\widetilde{\mathbf{x}}_{l,i}^{\rm v}$ is the input visual feature after the self-attention layer, $\mathbf{x}_{l+1,i}^{\rm v}$ is the output feature of the $l$-th visual Transformer block, $\mathrm{LN}(\cdot)$ denotes layer normalization, $\mathrm{MLP}(\cdot)$ is the original frozen FFN, and $\Delta\mathbf{x}_{l,i}^{\rm v}$ is the adapter residual. The adapter residual $\Delta\mathbf{x}_{l,i}^{\rm v}$ is computed via a bottleneck structure, consisting of a down-projection followed by a nonlinearity and an up-projection, as shown in Eq. (\ref{eq:cmsa_combined}).
\begin{equation}
    \begin{aligned}
        \mathbf{h}^{\rm v}_{l,i} &= \mathrm{ReLU}(\widetilde{\mathbf{x}}_{l,i}^{\rm v}{\mathbf{W}_{\rm down}^{\rm v}}^{\top} + \mathbf{b}_{\rm down}^{\rm v}), \\
        \Delta\mathbf{x}_{l,i}^{\rm v} &= s \cdot (\mathbf{h}^{\rm v}_{l,i}{\mathbf{W}_{\rm up}^{\rm v}}^{\top} + \mathbf{b}_{\rm up}^{\rm v}),
    \end{aligned}
    \label{eq:cmsa_combined}
\end{equation}
where $\mathbf{h}^{\rm v}_{l,i} \in \mathbb{R}^{\hat{d}_{\rm v}}$ is the intermediate bottleneck feature, $\mathbf{W}_{\rm down}^{\rm v} \in \mathbb{R}^{\hat{d}_{\rm v} \times d_{\rm v}}$ and $\mathbf{W}_{\rm up}^{\rm v} \in \mathbb{R}^{d_{\rm v} \times \hat{d}_{\rm v}}$ are learnable projection matrices, $\mathbf{b}_{\rm down}^{\rm v}$ and $\mathbf{b}_{\rm up}^{\rm v}$ are the corresponding bias terms, $s$ is a scaling factor, and $\hat{d}_{\rm v} < d_{\rm v}$ is the visual bottleneck dimension. Following the LoRA-style initialization, $\mathbf{W}_{\rm down}^{\rm v}$ is initialized with a Gaussian distribution while $\mathbf{W}_{\rm up}^{\rm v}$ and $\mathbf{b}_{\rm up}^{\rm v}$ are initialized to zero, so that the adapter residual $\Delta\mathbf{x}_{l,i}^{\rm v}$ is zero at the start of training. This preserves the pre-trained feature space at initialization and guarantees the stability of the SimGCD self-distillation process.

\emph{2) Visual Injection Adapter in the Text Branch.} For the text branch, we augment the basic adapter with a visual injection mechanism to bridge the cross-modal gap and enhance text representations with visual information. Specifically, we extract the \texttt{[CLS]} token $\mathbf{c}^{\rm v}_{l,i} \in \mathbb{R}^{d_{\rm v}}$ for $i$-th sample from the $l$-th layer of the visual encoder as the guidance signal. To enable parameter-efficient and stable cross-modal fusion without directly modifying the high-dimensional feature space, the cross-modal association is established within the low-dimensional bottleneck space, as shown in Eq. (\ref{eq:cmsa_down2}).
\begin{equation}
    \mathbf{h}^{\rm t}_{l,i} = {\rm ReLU}\Big(\underbrace{\tilde{\mathbf{x}}^{\rm t}_{l,i} \mathbf{W}^{\rm t\top}_{\rm down} + \mathbf{b}^{\rm t}_{\rm down}}_{\text{text feature projection}} + \underbrace{s_{\rm ctx}\, \mathbf{c}^{\rm v}_{l,i} \mathbf{W}^\top_{\rm cross}}_{\text{visual guidance injection}}\Big),
    \label{eq:cmsa_down2}
\end{equation}
where $\mathbf{h}^{\rm t}_{l,i} \in \mathbb{R}^{\hat{d}_{\rm t}}$ is the fused bottleneck feature combining text and visual information, $\widetilde{\mathbf{x}}_{l,i}^{\rm t}$ is the input text feature after the self-attention layer, $\mathbf{W}_{\rm down}^{\rm t} \in \mathbb{R}^{\hat{d}_{\rm t} \times d_{\rm t}}$ is a learnable down-projection matrix, $\mathbf{W}_{\rm cross} \in \mathbb{R}^{\hat{d}_{\rm t} \times d_{\rm v}}$ is a learnable cross-projection matrix that maps the visual guidance into the text bottleneck space, $s_{\rm ctx}$ is a scalar controlling the injection strength, and $\hat{d}_{\rm t} < d_{\rm t}$ is the text bottleneck dimension. The visual injection adapter residual is then obtained via Eq. (\ref{eq:cmsa_up2}).
\begin{equation}
    \Delta\mathbf{x}^{\rm t}_{l,i} = s \cdot \left(\mathbf{h}^{\rm t}_{l,i}{\mathbf{W}_{\rm up}^{\rm t}}^{\top} + \mathbf{b}_{\rm up}^{\rm t}\right),
    \label{eq:cmsa_up2}
\end{equation}
where $\mathbf{W}_{\rm up}^{\rm t} \in \mathbb{R}^{d_{\rm t} \times \hat{d}_{\rm t}}$ is a learnable up-projection matrix. The adapter parameters follow the same LoRA-style initialization as the visual branch. 
Through this layer-wise injection mechanism, the text branch progressively absorbs discriminative visual context across Transformer layers, achieving deep cross-modal synergy during encoding while keeping the visual branch fully independent for inference.

\subsection{Neighborhood Mutual Learning (NML)}
\label{sec:nml}

We propose Neighborhood Mutual Learning (NML), which exchanges information between modalities via local neighborhood structures instead of fixed global anchors. Specifically, NML first constructs modality-specific neighborhoods for each sample, and then exchanges their neighborhood distributions between the visual and text branches, enforcing consistency via bidirectional KL divergence.
 
\emph{1) Neighborhood Set Construction.}
We denote the batch feature matrices of the visual branch and the enhanced text branch as $\mathbf{Z}^{\rm v} \in \mathbb{R}^{|\mathcal{B}| \times d_{\rm v}}$ and $\mathbf{Z}^{\rm t} \in \mathbb{R}^{|\mathcal{B}| \times d_{\rm t}}$, respectively, where each row is the $\ell_2$-normalized feature of the sample. The corresponding pairwise cosine similarity matrices are computed as $\mathbf{S}^{\rm v} = \mathbf{Z}^{\rm v}(\mathbf{Z}^{\rm v})^{\top}$ and $\mathbf{S}^{\rm t} = \mathbf{Z}^{\rm t}(\mathbf{Z}^{\rm t})^{\top}$. To avoid trivial self-matching, we mask diagonal entries of the similarity matrices. The visual neighborhood index set $\mathcal{N}_i^{\rm v}$ of the $i$-th sample, which defines its local neighborhood in the visual feature space, is constructed by selecting the $K_{\rm nei}$ most similar samples, as formulated in Eq.~(\ref{eq:topk}).
\begin{equation}
    \mathcal{N}_i^{\rm v} = \left\{j \mid \mathbf{S}_{ij}^{\rm v} \in \mathrm{Top}\text{-}K(\mathbf{S}_{i,:}^{\rm v}, K_{\rm nei}) \right\}.
    \label{eq:topk}
\end{equation}
{\color{black}
For the text neighborhood index set $\mathcal{N}_i^{\rm t}$, it is defined analogously over $\mathbf{S}^{\rm t}$.

\emph{2) Bidirectional Neighborhood Distribution Mutual Learning.} 
For the $i$-th sample, the neighborhood distribution $\mathbf{q}_i^{\rm v}$ over $\mathcal{N}_i^{\rm v}$ quantifies the relative similarity to each neighbor in the visual feature space, where $\tau_{\rm nei}$ is a temperature hyper-parameter, as shown in Eq.~(\ref{eq:qprovisual}).
\begin{equation}
    \mathbf{q}_i^{\rm v}(j) = \frac{\exp(\mathbf{S}_{ij}^{\rm v} / \tau_{\rm nei})}{\sum_{k \in \mathcal{N}_i^{\rm v}} \exp(\mathbf{S}_{ik}^{\rm v} / \tau_{\rm nei})}, \quad j \in \mathcal{N}_i^{\rm v}.
    \label{eq:qprovisual}
\end{equation}
To encourage the enhanced text branch to preserve the same inter-sample relational structure as the visual branch, a corresponding distribution $\mathbf{q}_i^{\rm t}$ is computed using the text similarity matrix $\mathbf{S}^{\rm t}$ over the same index set $\mathcal{N}_i^{\rm v}$, as shown in Eq.~(\ref{eq:qprotext}).
\begin{equation}
    \mathbf{q}_i^{\rm t}(j) = \frac{\exp(\mathbf{S}_{ij}^{\rm t} / \tau_{\rm nei})}{\sum_{k \in \mathcal{N}_i^{\rm v}} \exp(\mathbf{S}_{ik}^{\rm t} / \tau_{\rm nei})}, \quad j \in \mathcal{N}_i^{\rm v}.
    \label{eq:qprotext}
\end{equation}
The visual-to-text mutual learning loss is defined as the KL divergence between $\mathbf{q}_i^{\rm v}$ and $\mathbf{q}_i^{\rm t}$, as shown in Eq.~(\ref{eq:lossv2t}).
\begin{equation}
    \mathcal{L}_{\rm v2t} = \frac{1}{|\mathcal{B}|} \sum_{i \in \mathcal{B}} \mathbb{D}_{\rm KL}(\mathbf{q}_i^{\rm v} \| \mathbf{q}_i^{\rm t}).
    \label{eq:lossv2t}
\end{equation}
Symmetrically, the text-to-visual loss $\mathcal{L}_{\rm t2v}$ is obtained by constructing the neighborhood $\mathcal{N}^{\rm t}_i$ over $\mathbf{S}^{\rm t}$ instead of $\mathbf{S}^{\rm v}$ and reversing the KL direction, following the same formulation as Eqs.~(\ref{eq:qprovisual})--(\ref{eq:lossv2t}). The total NML loss is formulated in Eq.~(\ref{eq:nml_total}).
\begin{equation}
    \mathcal{L}_{\rm nml} = \mathcal{L}_{\rm v2t} + \mathcal{L}_{\rm t2v}.
    \label{eq:nml_total}
\end{equation}
By replacing fixed global anchors with sample-adaptive local neighborhoods, NML enables fine-grained cross-modal information exchange and provides richer relational supervision across all samples regardless of their distance to any class center.

\emph{3) Overall Objective.}
The full SDBA framework is jointly optimized by integrating the intra-modal losses of both branches, the cross-modal mutual learning loss from the baseline, and the NML loss:
\begin{equation}
    \mathcal{L}_{\rm total} = \mathcal{L}_{\rm db}^{\rm v} + \mathcal{L}_{\rm db}^{\rm t} + \mathcal{L}_{\rm ml} + \lambda_{\rm nml}\mathcal{L}_{\rm nml},
    \label{eq:total_loss}
\end{equation}
where the first three terms inherit directly from the baseline framework (Sec.~\ref{sec:dualbranchframe}), and $\lambda_{\rm nml}$ is a balancing hyper-parameter. By inserting CMSA and NML into the baseline pipeline, SDBA functions as a plug-and-play enhancement compatible with existing dual-branch methods.

\begin{table}[bp]
\begin{center}
\renewcommand{\arraystretch}{1.5}
\setlength{\tabcolsep}{1pt}
\caption{Statistics of the benchmark datasets used in our experiments.}
\label{table:dataset_setting}
\scalebox{1}{
\begin{tabular}{cccccc}
\toprule
\multirow{2}{*}{\textbf{Benchmark}} & \multirow{2}{*}{\textbf{Dataset}} & \multicolumn{2}{c}{\textbf{Labeled}} & \multicolumn{2}{c}{\textbf{Unlabeled}}\\
\cmidrule(r){3-4} \cmidrule(r){5-6}
& & \textbf{\# Images} & \textbf{\# Classes} & \textbf{\# Images} & \textbf{\# Classes}\\
\midrule
\multirow{3}{*}{\makecell{Semantic Shift\\Benchmark}} 
& CUB-200~\cite{welinder2010caltech}        & 1.5K  & 100 & 4.5K  & 200 \\
& Aircraft~\cite{DBLP:journals/corr/MajiRKBV13}          & 1.7K  & 50  & 5.0K  & 100 \\
& Stanford Cars~\cite{krause20133d}     & 2.0K  & 98  & 6.1K  & 196 \\
\midrule
\multirow{3}{*}{\makecell{Generic Image\\Recognition\\Benchmark}}
& CIFAR-10~\cite{krizhevsky2009learning}  & 12.5K & 5   & 37.5K & 10  \\
& CIFAR-100~\cite{krizhevsky2009learning} & 20.0K & 80  & 30.0K & 100 \\
& ImageNet-100~\cite{DBLP:conf/eccv/TianKI20} & 31.9K & 50  & 95.3K & 100 \\
\bottomrule
\end{tabular}
}
\end{center}
\end{table}

\section{Experiments}
\label{section:experiment}
 
\subsection{Experimental Setup}

\begin{table*}[!htbp]
\renewcommand{\arraystretch}{1.5}
\setlength{\tabcolsep}{4pt}
\caption{\textbf{Evaluation results (\%) on the semantic shift benchmark (SSB) and the generic image recognition benchmark.} The best results are highlighted in \textbf{bold} and the second-best are \underline{underlined}. ``$-$'' indicates results not reported in the original paper.}
\label{tab:results_all}
\centering
\scalebox{0.9}{
\begin{threeparttable}
\begin{tabular}{cc|ccc|ccc|ccc|ccc|ccc|ccc}
\toprule
\multirow{3}{*}{\textbf{Method}} & \multirow{3}{*}{\textbf{Backbone}} & \multicolumn{9}{c|}{\textbf{Semantic Shift Benchmark}} & \multicolumn{9}{c}{\textbf{Generic Image Recognition Benchmark}} \\
\cmidrule(lr){3-11}\cmidrule(lr){12-20}
& & \multicolumn{3}{c|}{CUB-200} & \multicolumn{3}{c|}{Stanford Cars} & \multicolumn{3}{c|}{Aircraft} & \multicolumn{3}{c|}{CIFAR-10} & \multicolumn{3}{c|}{CIFAR-100} & \multicolumn{3}{c}{ImageNet-100} \\
\cmidrule(lr){3-5}\cmidrule(lr){6-8}\cmidrule(lr){9-11}\cmidrule(lr){12-14}\cmidrule(lr){15-17}\cmidrule(lr){18-20}
& & \cellcolor{blue!10}\textbf{All} & \textbf{Old} & \textbf{New} & \cellcolor{blue!10}\textbf{All} & \textbf{Old} & \textbf{New} & \cellcolor{blue!10}\textbf{All} & \textbf{Old} & \textbf{New} & \cellcolor{blue!10}\textbf{All} & \textbf{Old} & \textbf{New} & \cellcolor{blue!10}\textbf{All} & \textbf{Old} & \textbf{New} & \cellcolor{blue!10}\textbf{All} & \textbf{Old} & \textbf{New} \\
\midrule
$k$-means~\cite{DBLP:conf/soda/ArthurV07}       & DINO & \cellcolor{blue!10}34.3 & 38.9 & 32.1 & \cellcolor{blue!10}12.8 & 10.6 & 13.8 & \cellcolor{blue!10}16.0 & 14.4 & 16.8 & \cellcolor{blue!10}83.6 & 85.7 & 82.5 & \cellcolor{blue!10}52.0 & 52.2 & 50.8 & \cellcolor{blue!10}72.7 & 75.5 & 71.3 \\
RS+~\cite{DBLP:journals/pami/HanREVZ22}          & DINO & \cellcolor{blue!10}33.3 & 51.6 & 24.2 & \cellcolor{blue!10}28.3 & 61.8 & 12.1 & \cellcolor{blue!10}26.9 & 36.4 & 22.2 & \cellcolor{blue!10}46.8 & 19.2 & 60.5 & \cellcolor{blue!10}58.2 & 77.6 & 19.3 & \cellcolor{blue!10}37.1 & 61.6 & 24.8 \\
UNO+~\cite{DBLP:conf/iccv/FiniSLZN021}           & DINO & \cellcolor{blue!10}35.1 & 49.0 & 28.1 & \cellcolor{blue!10}35.5 & 70.5 & 18.6 & \cellcolor{blue!10}40.3 & 56.4 & 32.2 & \cellcolor{blue!10}68.6 & \textbf{98.3} & 53.8 & \cellcolor{blue!10}69.5 & 80.6 & 47.2 & \cellcolor{blue!10}70.3 & 95.0 & 57.9 \\
ORCA~\cite{DBLP:conf/iclr/CaoBL22}               & DINO & \cellcolor{blue!10}35.3 & 45.6 & 30.2 & \cellcolor{blue!10}23.5 & 50.1 & 10.7 & \cellcolor{blue!10}22.0 & 31.8 & 17.1 & \cellcolor{blue!10}81.8 & 86.2 & 79.6 & \cellcolor{blue!10}69.0 & 77.4 & 52.0 & \cellcolor{blue!10}73.5 & 92.6 & 63.9 \\
GCD~\cite{DBLP:conf/cvpr/gcdVazeHVZ22}           & DINO & \cellcolor{blue!10}51.3 & 56.6 & 48.7 & \cellcolor{blue!10}39.0 & 57.6 & 29.9 & \cellcolor{blue!10}45.0 & 41.1 & 46.9 & \cellcolor{blue!10}91.5 & 97.9 & 88.2 & \cellcolor{blue!10}73.0 & 76.2 & 66.5 & \cellcolor{blue!10}74.1 & 89.8 & 66.3 \\
DCCL~\cite{DBLP:conf/cvpr/PuZS23}                & DINO & \cellcolor{blue!10}63.5 & 60.8 & 64.9 & \cellcolor{blue!10}43.1 & 55.7 & 36.2 & \cellcolor{blue!10}- & - & - & \cellcolor{blue!10}96.3 & 96.5 & 96.9 & \cellcolor{blue!10}75.3 & 76.8 & 70.2 & \cellcolor{blue!10}80.5 & 90.5 & 76.2 \\
PromptCAL~\cite{DBLP:conf/cvpr/promptcalgcdZhangKSNCK23} & DINO & \cellcolor{blue!10}62.9 & 64.4 & 62.1 & \cellcolor{blue!10}50.2 & 70.1 & 40.6 & \cellcolor{blue!10}52.2 & 52.2 & 52.3 & \cellcolor{blue!10}97.9 & 96.6 & 98.5 & \cellcolor{blue!10}81.2 & 84.2 & 75.3 & \cellcolor{blue!10}83.1 & 92.7 & 78.3 \\
SimGCD~\cite{DBLP:conf/iccv/simgcdWenZQ23}       & DINO & \cellcolor{blue!10}60.3 & 65.6 & 57.7 & \cellcolor{blue!10}53.8 & 71.9 & 45.0 & \cellcolor{blue!10}54.2 & 59.1 & 51.8 & \cellcolor{blue!10}97.1 & 95.1 & 98.1 & \cellcolor{blue!10}80.1 & 81.2 & 77.8 & \cellcolor{blue!10}83.0 & 93.1 & 77.9 \\
$\mu$GCD~\cite{DBLP:conf/nips/VazeVZ23}          & DINO & \cellcolor{blue!10}65.7 & 68.0 & 64.6 & \cellcolor{blue!10}56.5 & 68.1 & 50.9 & \cellcolor{blue!10}53.8 & 55.4 & 53.0 & \cellcolor{blue!10}- & - & - & \cellcolor{blue!10}- & - & - & \cellcolor{blue!10}- & - & - \\
LegoGCD~\cite{DBLP:conf/cvpr/CaoZWYS0L024}       & DINO & \cellcolor{blue!10}63.8 & 71.9 & 59.8 & \cellcolor{blue!10}57.3 & 75.7 & 48.4 & \cellcolor{blue!10}55.0 & 61.5 & 51.7 & \cellcolor{blue!10}97.1 & 94.3 & 98.5 & \cellcolor{blue!10}81.8 & 81.4 & 82.5 & \cellcolor{blue!10}86.3 & 94.5 & 82.1 \\
AdaptGCD~\cite{11142371}                         & DINO & \cellcolor{blue!10}68.8 & 74.5 & 65.9 & \cellcolor{blue!10}62.7 & 80.6 & 54.0 & \cellcolor{blue!10}57.9 & \underline{65.2} & 54.3 & \cellcolor{blue!10}97.9 & 95.5 & \textbf{99.0} & \cellcolor{blue!10}84.0 & 86.2 & 79.7 & \cellcolor{blue!10}86.4 & 94.9 & 82.1 \\
PartGCD~\cite{wang2026learning}                 & DINO & \cellcolor{blue!10}68.6 & 68.9 & 68.4 & \cellcolor{blue!10}65.6 & 79.5 & 58.9 & \cellcolor{blue!10}59.4 & 63.5 & 57.4 & \cellcolor{blue!10}97.3 & 95.7 & 98.1 & \cellcolor{blue!10}82.4 & 84.1 & 79.0 & \cellcolor{blue!10}85.8 & 94.4 & 81.5 \\
SADA~\cite{peng2026sharpness}                   & DINO & \cellcolor{blue!10}69.0 & 69.1 & 68.9 & \cellcolor{blue!10}65.5 & 79.7 & 58.8 & \cellcolor{blue!10}55.6 & 59.2 & 53.8 & \cellcolor{blue!10}- & - & - & \cellcolor{blue!10}83.7 & 84.6 & 82.0 & \cellcolor{blue!10}87.1 & 93.9 & 83.7 \\
\midrule
ClipGCD~\cite{DBLP:journals/corr/abs-2305-10420} & CLIP & \cellcolor{blue!10}62.8 & 77.1 & 55.7 & \cellcolor{blue!10}70.6 & 88.2 & 62.2 & \cellcolor{blue!10}50.0 & 56.6 & 46.5 & \cellcolor{blue!10}96.6 & 97.2 & 96.4 & \cellcolor{blue!10}85.2 & 85.0 & \textbf{85.6} & \cellcolor{blue!10}84.0 & 95.5 & 78.2 \\
GCD-CLIP~\cite{DBLP:conf/cvpr/gcdVazeHVZ22}          & CLIP & \cellcolor{blue!10}57.6 & 65.2 & 53.8 & \cellcolor{blue!10}65.1 & 75.9 & 59.8 & \cellcolor{blue!10}45.3 & 44.4 & 45.8 & \cellcolor{blue!10}94.0 & 97.3 & 92.3 & \cellcolor{blue!10}74.8 & 79.8 & 64.6 & \cellcolor{blue!10}75.8 & 87.3 & 70.0 \\
SimGCD-CLIP~\cite{DBLP:conf/iccv/simgcdWenZQ23}  & CLIP & \cellcolor{blue!10}71.7 & 76.5 & 69.4 & \cellcolor{blue!10}70.0 & 83.4 & 63.5 & \cellcolor{blue!10}54.3 & 58.4 & 52.2 & \cellcolor{blue!10}97.0 & 94.2 & 98.4 & \cellcolor{blue!10}81.1 & 85.0 & 73.3 & \cellcolor{blue!10}90.8 & 95.5 & 88.5 \\
GET~\cite{DBLP:conf/cvpr/WangPXYLC25}             & CLIP & \cellcolor{blue!10}77.0 & 78.1 & 76.4 & \cellcolor{blue!10}78.5 & 86.8 & 74.5 & \cellcolor{blue!10}\underline{58.9} & 59.6 & \underline{58.5} & \cellcolor{blue!10}97.2 & 94.6 & 98.5 & \cellcolor{blue!10}82.1 & 85.5 & 75.5 & \cellcolor{blue!10}\underline{91.7} & \underline{95.7} & \underline{89.7} \\
TextGCD~\cite{DBLP:conf/eccv/ZhengPLSZ24}        & CLIP & \cellcolor{blue!10}76.6 & \underline{80.6} & 74.7 & \cellcolor{blue!10}\underline{86.9} & 87.4 & \textbf{86.7} & \cellcolor{blue!10}- & - & - & \cellcolor{blue!10}98.2 & \underline{98.0} & 98.6 & \cellcolor{blue!10}85.7 & 86.3 & 84.6 & \cellcolor{blue!10}88.0 & 92.4 & 85.2 \\
\midrule
\rowcolor{blue!10}\textbf{SDBA + GET}     & CLIP & \cellcolor{blue!10}\textbf{79.4} & \textbf{81.6} & \textbf{78.3} & \cellcolor{blue!10}80.5 & \underline{87.8} & 77.0 & \cellcolor{blue!10}\textbf{67.1} & \textbf{71.2} & \textbf{65.1} & \cellcolor{blue!10}\underline{98.3} & 97.5 & \underline{98.7} & \cellcolor{blue!10}\underline{86.1} & \textbf{89.0} & 80.3 & \cellcolor{blue!10}\textbf{91.9} & \textbf{96.0} & \textbf{90.0} \\
\rowcolor{blue!10}\textbf{SDBA + TextGCD} & CLIP & \cellcolor{blue!10}\underline{77.6} & 78.0 & \underline{77.4} & \cellcolor{blue!10}\textbf{87.6} & \textbf{92.6} & \underline{85.1} & \cellcolor{blue!10}- & - & - & \cellcolor{blue!10}\textbf{98.5} & \textbf{98.3} & 98.6 & \cellcolor{blue!10}\textbf{87.1} & \underline{87.9} & \underline{85.4} & \cellcolor{blue!10}- & - & - \\
\bottomrule
\end{tabular}
\textsuperscript{\dag} TextGCD does not provide the retrieval lexicon for Aircraft, and ImageNet-100 experiments with TextGCD are omitted due to the substantial computational cost; results are therefore not reported.
\end{threeparttable}
}
\end{table*}

\emph{1) Datasets.}
We evaluate the proposed framework on six datasets across two benchmarks. The Semantic Shift Benchmark (SSB) comprises three fine-grained datasets: CUB-200, Aircraft, and Stanford Cars, which are particularly challenging due to high inter-class visual similarity. The Generic Image Recognition Benchmark includes CIFAR-10, CIFAR-100~\cite{krizhevsky2009learning}, and ImageNet-100~\cite{DBLP:conf/eccv/TianKI20}. Following the standard GCD protocol~\cite{DBLP:conf/cvpr/gcdVazeHVZ22}, a subset of classes is designated as old classes (80\% for CIFAR-100 and 50\% for all other datasets). Among the images of old classes, 50\% are used as the labeled set $\mathcal{D}_l$, while the remaining old-class images and all new-class images form the unlabeled set $\mathcal{D}_u$. Dataset statistics are provided in Table~\ref{table:dataset_setting}.

\emph{2) Implementation Details.}
We use the CLIP~\cite{DBLP:conf/icml/RadfordKHRGASAM21} pre-trained model as the backbone, with the visual branch employing the ViT-B/16~\cite{DBLP:conf/iclr/DosovitskiyB0WZ21} visual encoder and the text branch employing the corresponding CLIP text encoder. Unless otherwise specified, all experiments are conducted under the GET-based framework. SDBA is integrated into the respective baseline following the staged training strategy described in the Appendix B. For all datasets, adapters are inserted from layer 3 to the final layer, following common practice in adapter-based tuning. The key hyperparameters are listed in the Appendix C. The number of trainable parameters introduced by SDBA is comparable to that of the GET baseline, with detailed parameter counts and a fairness analysis provided in the Appendix D. In addition, following the architectural design of the underlying baseline, the GET-based variant shares a classification head across the two branches {and reports the visual-branch prediction results at inference}, while the TextGCD-based variant uses separate heads {and reports the fused prediction result}.

\emph{3) Evaluation.}
Following~\cite{DBLP:conf/cvpr/gcdVazeHVZ22}, we adopt clustering accuracy as the evaluation metric, measured separately over old classes, new classes, and all classes. For sample $\mathbf{x}_i^{\rm v}$ with ground truth label $y_i$ and predicted cluster assignment $\hat{y}_i$, the metrics $acc^{old}$, $acc^{new}$, and $acc^{all}$ are computed as:
\begin{equation}
\label{cal::metrics}
    \begin{aligned}
        acc^{old}=& \frac{1}{|\mathcal{D}^{u,old}|}\sum_{i=1}^{|\mathcal{D}^{u,old}|}\mathbbm{1}(y_i={\rm Permute}(\hat{y}_i))\\
        acc^{new}=& \frac{1}{|\mathcal{D}^{u,new}|}\sum_{i=1}^{|\mathcal{D}^{u,new}|}\mathbbm{1}(y_i={\rm Permute}(\hat{y}_i))\\
        acc^{all}=& \frac{1}{|\mathcal{D}^u|}\sum_{i=1}^{|\mathcal{D}^u|}\mathbbm{1}(y_i={\rm Permute}(\hat{y}_i))
    \end{aligned}
\end{equation}
where $\mathcal{D}^{u,old}$ and $\mathcal{D}^{u,new}$ denote the old-class and new-class subsets of the unlabeled set $\mathcal{D}^{u}$, and ${\rm Permute}(\cdot)$ denotes the optimal assignment between predicted cluster indices and ground truth labels via Hungarian matching.

\subsection{Comparison with State-of-the-Art}

We compare SDBA against a wide range of unimodal GCD methods, including $k$-means~\cite{DBLP:conf/soda/ArthurV07}, RS+~\cite{DBLP:journals/pami/HanREVZ22}, UNO+~\cite{DBLP:conf/iccv/FiniSLZN021}, and DINO-based baselines such as ORCA~\cite{DBLP:conf/iclr/CaoBL22}, GCD~\cite{DBLP:conf/cvpr/gcdVazeHVZ22}, DCCL~\cite{DBLP:conf/cvpr/PuZS23}, SimGCD~\cite{DBLP:conf/iccv/simgcdWenZQ23}, PromptCAL~\cite{DBLP:conf/cvpr/promptcalgcdZhangKSNCK23}, $\mu$GCD~\cite{DBLP:conf/nips/VazeVZ23}, LegoGCD~\cite{DBLP:conf/cvpr/CaoZWYS0L024}, AdaptGCD~\cite{11142371}, SADA\cite{peng2026sharpness}, and PartGCD\cite{wang2026learning}. Beyond these unimodal benchmarks, we further consider CLIP-based variants (e.g., SimGCD-CLIP~\cite{DBLP:conf/iccv/simgcdWenZQ23}, GCD-CLIP~\cite{DBLP:conf/cvpr/gcdVazeHVZ22}) and state-of-the-art multi-modal paradigms, including ClipGCD~\cite{DBLP:journals/corr/abs-2305-10420}, GET~\cite{DBLP:conf/cvpr/WangPXYLC25}, and TextGCD~\cite{DBLP:conf/eccv/ZhengPLSZ24}. These extensive comparisons establish a solid foundation for evaluating the effectiveness of our proposed framework.

\emph{1) Evaluation on the Semantic Shift Benchmark.}
The semantic shift benchmark datasets are particularly challenging owing to subtle inter-class visual differences. As shown in Table~\ref{tab:results_all}, SDBA consistently outperforms existing approaches across all tasks. By incorporating complementary modality information, it surpasses unimodal DINO-based methods; simultaneously, its superior cross-modal synergy enables it to outperform CLIP-based variants and existing multi-modal dual-branch frameworks. Specifically, SDBA + GET achieves significant gains of 2.4\%, 2.0\%, and 8.2\% on the ``All'' metric for CUB-200, SCars, and Aircraft, respectively. The improvement on these datasets validates that visual injection effectively suppresses the noise inherent in synthesized text embeddings, thereby enabling deeper cross-modal synergy between the two branches. SDBA + TextGCD also yields stable gains, confirming the scalability of our framework across both synthesis-based and retrieval-based text paradigms.

\emph{2) Evaluation on the Generic Image Recognition Benchmark.}
As shown in Table~\ref{tab:results_all}, SDBA yields stable performance gains on generic benchmarks. Specifically, SDBA + GET improves the ``All'' accuracy over GET by 1.1\%, 4.0\%, and 0.2\% on CIFAR-10, CIFAR-100, and ImageNet-100, respectively. These results demonstrate that the proposed framework effectively optimizes the joint representation space through cross-modal synergy. In conclusion, across varying datasets, SDBA maintains consistent effectiveness.

\subsection{Ablation Study}
 
\begin{table*}[!htbp]
\renewcommand{\arraystretch}{1.5}
\setlength{\tabcolsep}{7pt}
\centering
\caption{\textbf{Ablation study of SDBA components on Aircraft and CIFAR-100.} ``Basic Adapter'' refers to inserting standard adapters into both branches independently; ``Visual Injection'' refers to upgrading the text branch adapter to the Visual Injection Adapter of CMSA.}
\label{tab:ablation_study}
\scalebox{1}{
\begin{threeparttable}
\begin{tabular}{l c c c  cccccccccccc}
\toprule
\multirow{3}{*}{\textbf{ID}} &
\multicolumn{2}{c}{\textbf{CMSA}} &
\multirow{3}{*}{\makecell{\textbf{NML}}} &
\multicolumn{6}{c}{{Aircraft}} &
\multicolumn{6}{c}{{CIFAR-100}} \\
\cmidrule(lr){2-3}  \cmidrule(lr){5-10} \cmidrule(l){11-16}
& \multirow{2}{*}{\makecell{\textbf{Basic}\\\textbf{Adapter}}} &
\multirow{2}{*}{\makecell{\textbf{Visual}\\\textbf{Injection}}} & &
\multicolumn{3}{c}{\textbf{Visual}} &
\multicolumn{3}{c}{\textbf{Text}} &
\multicolumn{3}{c}{\textbf{Visual}} &
\multicolumn{3}{c}{\textbf{Text}} \\
\cmidrule(lr){5-7} \cmidrule(lr){8-10} \cmidrule(lr){11-13} \cmidrule(l){14-16}
& & & &
\cellcolor{blue!10}\textbf{All} & \textbf{Old} & \textbf{New} & \cellcolor{blue!10}\textbf{All} & \textbf{Old} & \textbf{New} &
\cellcolor{blue!10}\textbf{All} & \textbf{Old} & \textbf{New} & \cellcolor{blue!10}\textbf{All} & \textbf{Old} & \textbf{New} \\
\midrule
\tt{(1)} & \XSolidBrush & \XSolidBrush & \XSolidBrush &
\cellcolor{blue!10}58.4 & 59.3 & 58.0 & \cellcolor{blue!10}46.5 & 53.1 & 43.2 &
\cellcolor{blue!10}82.1 & 85.7 & 75.0 & \cellcolor{blue!10}77.3 & 81.8 & 68.1 \\
\tt{(2)} & \Checkmark & \XSolidBrush & \XSolidBrush &
\cellcolor{blue!10}62.4 & 68.1 & 59.5 & \cellcolor{blue!10}49.5 & 54.6 & 46.9 &
\cellcolor{blue!10}83.8 & \textbf{89.2} & 73.0 & \cellcolor{blue!10}75.9 & 81.5 & 64.7 \\
\tt{(3)} & \Checkmark & \Checkmark & \XSolidBrush &
\cellcolor{blue!10}64.9 & \textbf{71.3} & 61.7 & \cellcolor{blue!10}64.9 & 71.1 & 61.8 &
\cellcolor{blue!10}86.0 & 89.0 & 79.9 & \cellcolor{blue!10}86.0 & 89.0 & 79.9 \\
\rowcolor{blue!10}\tt{(4)} &  \Checkmark &  \Checkmark & \Checkmark &
\cellcolor{blue!10}\textbf{67.1} & {71.2} & \textbf{65.1} & \cellcolor{blue!10}\textbf{67.1} & \textbf{71.2} & \textbf{65.0} &
\cellcolor{blue!10}\textbf{86.1} & {89.0} & \textbf{80.3} & \cellcolor{blue!10}\textbf{86.1} & \textbf{89.1} & \textbf{80.1} \\
\bottomrule
\end{tabular}
\textsuperscript{\dag} As the original paper does not report the complete per-branch performance, the GET baseline (ID {\tt (1)}) corresponds to our reproduced results.
\end{threeparttable}
}
\end{table*}

Ablation experiments are conducted on Aircraft and CIFAR-100 using GET as the base framework, with results reported in Table~\ref{tab:ablation_study}. Extended ablation results on some other datasets are provided in the Appendix D.

\emph{1) Effect of Cross-Modal Synergistic Adapter.}
Introducing the basic adapter (ID~{\tt{(2)}}) generally improves the performance over the baseline (ID~{\tt{(1)}}), laying a stronger feature foundation for subsequent cross-modal interaction. Upon further introducing visual injection (ID~{\tt{(3)}}), performance improves substantially, particularly on Aircraft, where the text branch accuracy achieves a large gain and the performance gap between the two branches is markedly reduced. This is consistent with the motivation of CMSA: the text branch suffers from the bias and noise in the derived texts or text embeddings when encoded independently, and visual injection mitigates this during encoding. Fig.~\ref{fig:training_plots} further illustrates this effect. Without visual injection, the visual branch converges to a significantly higher level than the text branch, with a persistent gap throughout training. With visual injection, both branches rise synchronously and converge at a comparable level, confirming that visual injection effectively enhances text branch learning and enables overall reinforcement.

\begin{figure}[t]
  \centering
   \includegraphics[width=0.95\linewidth]{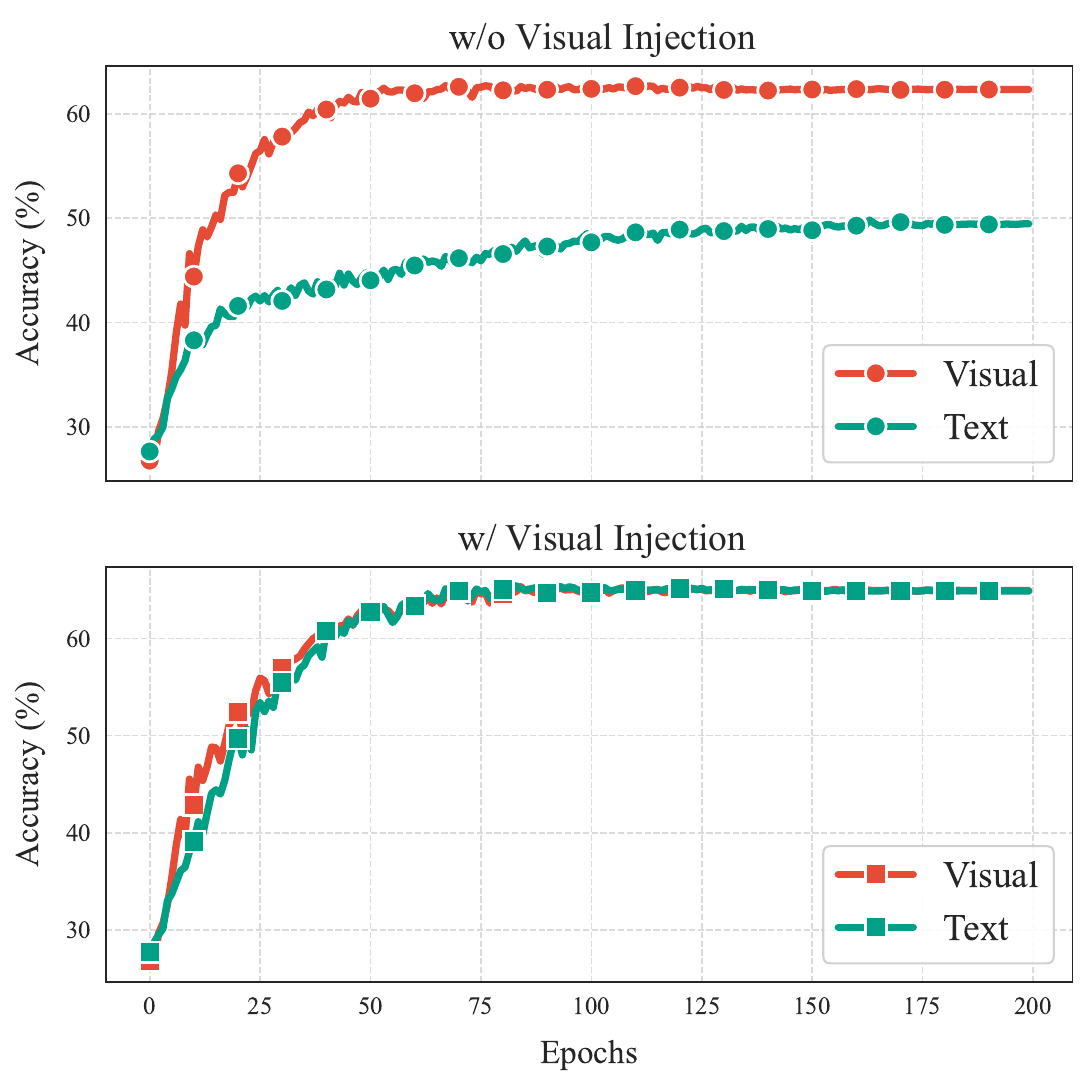}
   \vspace{-2mm}
   \caption{Training accuracy curves of the visual and text branches on Aircraft, without visual injection (top) and with visual injection (bottom).
   }
   \label{fig:training_plots}
\end{figure}
 
\emph{2) Effect of Neighborhood Mutual Learning.}
Building on CMSA, the introduction of NML (ID~{\tt{(4)}}) yields further gains in overall accuracy across both datasets, with improvements more pronounced on new-class recognition. NML replaces fixed global anchors with local sample neighborhoods, enabling the model to capture fine-grained inter-sample relational structure that global anchors fail to reflect. In contrast, the baseline GET employs CICO, which is driven by old-class prototypes and thus provides limited supervision for new-class samples. By alleviating this bias toward old classes, NML leads to more substantial performance gains on new categories.

\subsection{Analysis Experiments}
\label{sec:analysis}
 
\emph{1) Sensitivity to Visual Injection Intensity.}
The parameter $s_{\rm ctx}$ controls the injection strength of visual context into the text branch. We conduct sensitivity analysis on Aircraft and SCars over different values of $s_{\rm ctx}$, as illustrated in Fig.~\ref{fig:lambda_ctx_ablation}. As $s_{\rm ctx}$ increases, performance on both branches initially improves, confirming that visual context injection effectively enhances the discriminative quality of both branches. However, once $s_{\rm ctx}$ exceeds an optimal threshold (approximately 0.5 for Aircraft and 0.3 for SCars), performance declines for both branches. 
This indicates that excessive visual injection overwhelms the semantic information from the text branch, degrading the overall performance.

\begin{figure}[t]
  \centering
   \includegraphics[width=1.0\linewidth]{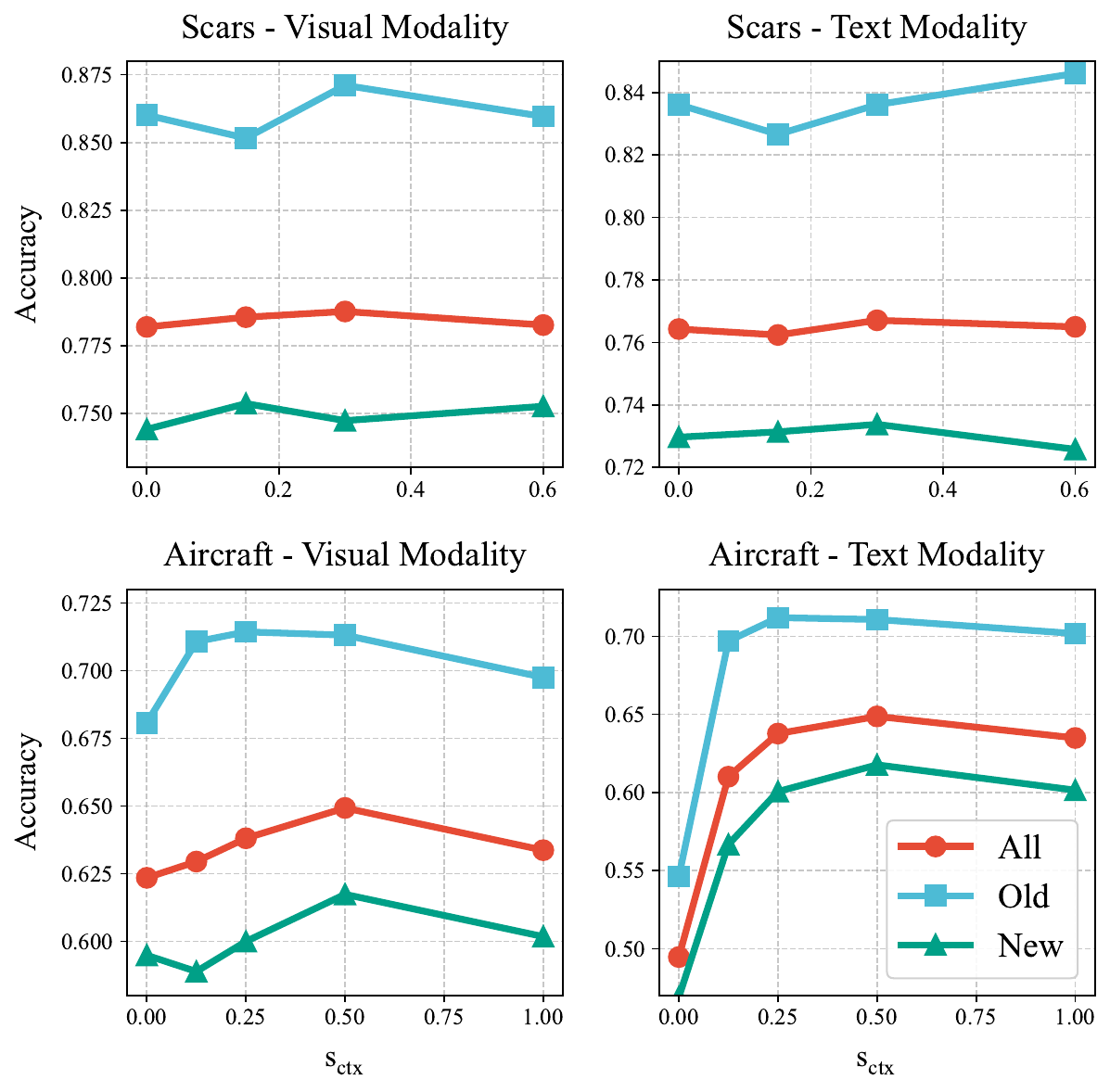}
   \vspace{-2mm}
   \caption{Accuracy of the visual and text branches on the All, Old, and New classes with different $s_{\rm ctx}$, evaluated on SCars (top row) and Aircraft (bottom row).}
   \label{fig:lambda_ctx_ablation}
\end{figure}
 
\emph{2) Impact of Neighborhood Size.}
The neighborhood size $K_{\rm nei}$ determines the scope of the local constraint in NML. We study the effect of $K_{\rm nei}$ on Aircraft and SCars with values ranging over 5, 10, 20, 30, and 40, as shown in Table~\ref{tab:topk_ablation}. Performance first increases then decreases on both datasets. A small $K_{\rm nei}$ (e.g., 5) is insufficient to capture the local relational structure, limiting alignment effectiveness. Conversely, an excessively large $K_{\rm nei}$ incorporates semantically distant samples into the neighborhood, introducing noisy relational signals that degrade the quality of the neighborhood distribution. The optimal $K_{\rm nei}$ is 30 for Aircraft and 20 for SCars.

\begin{table}[t]
\renewcommand{\arraystretch}{1.3}
\setlength{\tabcolsep}{5.5pt}
\caption{Effect of neighborhood size $K_{\rm nei}$ on the visual and text branch clustering accuracy (All/Old/New) on Aircraft and SCars.
}
\label{tab:topk_ablation}
\centering
\scalebox{1}{
\begin{tabular}{cccccccc}
\toprule
\multirow{2}{*}{\textbf{Dataset}} & \multirow{2}{*}{$K_{\rm nei}$} & \multicolumn{3}{c}{\textbf{Visual Branch}} & \multicolumn{3}{c}{\textbf{Text Branch}} \\
\cmidrule(lr){3-5} \cmidrule(lr){6-8}
& & \cellcolor{blue!10}\textbf{All} & \textbf{Old} & \textbf{New} & \cellcolor{blue!10}\textbf{All} & \textbf{Old} & \textbf{New} \\
\midrule
\multirow{5}{*}{Aircraft}
& 5  & \cellcolor{blue!10}64.8 & 70.8 & 61.8 & \cellcolor{blue!10}64.7 & 70.6 & 61.7 \\
& 10 & \cellcolor{blue!10}66.4 & \textbf{74.1} & 62.6 & \cellcolor{blue!10}66.2 & \textbf{73.8} & 62.3 \\
& 20 & \cellcolor{blue!10}65.3 & 70.7 & 62.6 & \cellcolor{blue!10}65.2 & 70.8 & 62.4 \\
& \cellcolor{blue!10}30 & \cellcolor{blue!10}\textbf{67.1} & \cellcolor{blue!10}71.2 & \cellcolor{blue!10}\textbf{65.1} & \cellcolor{blue!10}\textbf{67.1} & \cellcolor{blue!10}71.2 & \cellcolor{blue!10}\textbf{65.0} \\
& 40 & \cellcolor{blue!10}65.7 & 72.2 & 62.5 & \cellcolor{blue!10}65.7 & 72.2 & 62.5 \\
\midrule
\multirow{5}{*}{SCars}
& 5  & \cellcolor{blue!10}79.8 & 86.4 & 76.7 & \cellcolor{blue!10}77.6 & 82.9 & 75.0 \\
& 10 & \cellcolor{blue!10}79.9 & 86.3 & 76.8 & \cellcolor{blue!10}77.7 & 82.9 & 75.2 \\
& \cellcolor{blue!10}20 & \cellcolor{blue!10}\textbf{80.5} & \cellcolor{blue!10}\textbf{87.8} & \cellcolor{blue!10}\textbf{77.0} & \cellcolor{blue!10}\textbf{78.2} & \cellcolor{blue!10}84.3 & \cellcolor{blue!10}\textbf{75.3} \\
& 30 & \cellcolor{blue!10}80.2 & 87.7 & 76.6 & \cellcolor{blue!10}78.0 & 84.2 & 75.0 \\
& 40 & \cellcolor{blue!10}79.9 & 87.0 & 76.4 & \cellcolor{blue!10}77.3 & \textbf{84.8} & 73.7 \\
\bottomrule
\end{tabular}
}
\end{table}
 
\emph{3) Impact of NML Loss Weight.}
$\lambda_{\rm nml}$ controls the strength of the neighborhood consistency constraint. We study different values of $\lambda_{\rm nml}$ on Aircraft and SCars, with results reported in Table~\ref{tab:lambda_nml_ablation}. On both datasets, performance first increases then decreases as $\lambda_{\rm nml}$ grows, with optimal values of 1 for Aircraft and 0.1 for SCars. When $\lambda_{\rm nml}$ is too small, the neighborhood constraint is too weak to provide effective relational supervision across modalities. Conversely, an excessively large $\lambda_{\rm nml}$ enforces excessive consistency between the two branches, reducing the complementarity of the two modalities and leading to performance degradation.
 
\begin{table}[t]
\renewcommand{\arraystretch}{1.3}
\setlength{\tabcolsep}{5.5pt}
\caption{Effect of NML loss weight $\lambda_{\rm nml}$ on the visual and text branch clustering accuracy (All/Old/New) on Aircraft and SCars.
}
\label{tab:lambda_nml_ablation}
\centering
\scalebox{1}{
\begin{tabular}{cccccccc}
\toprule
\multirow{2}{*}{\textbf{Dataset}} & \multirow{2}{*}{$\lambda_{\rm nml}$} & \multicolumn{3}{c}{\textbf{Visual Branch}} & \multicolumn{3}{c}{\textbf{Text Branch}} \\
\cmidrule(lr){3-5} \cmidrule(lr){6-8}
& & \cellcolor{blue!10}\textbf{All} & \textbf{Old} & \textbf{New} & \cellcolor{blue!10}\textbf{All} & \textbf{Old} & \textbf{New} \\
\midrule
\multirow{4}{*}{Aircraft}
& 0.25 & \cellcolor{blue!10}64.3 & 70.8 & 61.1 & \cellcolor{blue!10}64.6 & \textbf{72.7} & 60.6 \\
& 0.5  & \cellcolor{blue!10}64.9 & \textbf{72.5} & 61.1 & \cellcolor{blue!10}64.5 & 72.3 & 60.6 \\
& \cellcolor{blue!10}1    & \cellcolor{blue!10}\textbf{67.1} & \cellcolor{blue!10}71.2 & \cellcolor{blue!10}\textbf{65.1} & \cellcolor{blue!10}\textbf{67.1} & \cellcolor{blue!10}71.2 & \cellcolor{blue!10}\textbf{65.0} \\
& 2    & \cellcolor{blue!10}67.0 & 71.4 & 64.8 & \cellcolor{blue!10}67.0 & 71.4 & 64.8 \\
\midrule
\multirow{4}{*}{SCars}& 0.05	& \cellcolor{blue!10}80.0 &  85.2 & \textbf{77.5} & \cellcolor{blue!10}77.7 & 82.8 & 75.2 \\
& \cellcolor{blue!10}0.1 & \cellcolor{blue!10}\textbf{80.5} & \cellcolor{blue!10}\textbf{87.8} & \cellcolor{blue!10}{77.0} & \cellcolor{blue!10}\textbf{78.2} & \cellcolor{blue!10}84.3 & \cellcolor{blue!10}\textbf{75.3} \\
& 0.2 & \cellcolor{blue!10}79.2 & 85.9 & 75.9 & \cellcolor{blue!10}77.1 & 84.4 & 73.6\\
& 0.4 & \cellcolor{blue!10}78.5 & 87.1 & 74.3 & \cellcolor{blue!10}77.1 & \textbf{85.6} & 73.0\\
\bottomrule
\end{tabular}
}
\end{table}
 
\emph{4) Comparison of Different Fusion Strategies.}
To validate the design of CMSA, we compare it against three alternative fusion baselines: Cross Attention, which fuses the last-layer visual tokens into the last-layer text tokens via a cross attention mechanism; Concat Linear, which concatenates visual and text features followed by a linear projection; and CoCoOp~\cite{Zhou2022CoCoOp}, which dynamically generates instance-conditioned prompts from visual features. Results on Aircraft are reported in Table~\ref{tab:fusion_method_comparison}. CMSA consistently outperforms all three baselines, benefiting from its preservation of the pre-trained feature space at initialization, which better accommodates the self-distillation paradigm of SimGCD. This complements the finding in GET~\cite{DBLP:conf/cvpr/WangPXYLC25} that naive fusion underperforms dual-branch learning, while well-designed fusion like CMSA still brings further gains.

\begin{table}[t]
\renewcommand{\arraystretch}{1.3}
\setlength{\tabcolsep}{5pt}
\caption{Comparison of fusion strategies on Aircraft. NML is excluded from all settings for fair comparison.}
\label{tab:fusion_method_comparison}
\centering
\scalebox{1}{
\begin{tabular}{lcccccc}
\toprule
\multirow{2}{*}{\textbf{Method}} & \multicolumn{3}{c}{\textbf{Visual Branch}} & \multicolumn{3}{c}{\textbf{Text Branch}} \\
\cmidrule(lr){2-4} \cmidrule(lr){5-7}
& \cellcolor{blue!10}\textbf{All} & \textbf{Old} & \textbf{New} & \cellcolor{blue!10}\textbf{All} & \textbf{Old} & \textbf{New} \\
\midrule
Cross Attention      & \cellcolor{blue!10}56.0 & 59.4 & 54.3 & \cellcolor{blue!10}56.0 & 59.4 & 54.3 \\
Concat Linear        & \cellcolor{blue!10}57.5 & 61.3 & 55.6 & \cellcolor{blue!10}57.4 & 61.9 & 55.1 \\
CoCoOp               & \cellcolor{blue!10}59.9 & 62.4 & 58.6 & \cellcolor{blue!10}60.2 & 62.2 & 59.1 \\
\cellcolor{blue!10}\textbf{Ours (CMSA)} & \cellcolor{blue!10}\textbf{64.9} & \cellcolor{blue!10}\textbf{71.3} & \cellcolor{blue!10}\textbf{61.7} & \cellcolor{blue!10}\textbf{64.9} & \cellcolor{blue!10}\textbf{71.1} & \cellcolor{blue!10}\textbf{61.8} \\
\bottomrule
\end{tabular}
}
\end{table}

\emph{5) Feature Space Visualization.}
To more intuitively assess the impact of visual injection on the features of both branches, t-SNE visualization is conducted on the Aircraft dataset as shown in Fig.~\ref{fig:tsne}. Without the injection mechanism (top row), the text branch features exhibit dispersed and overlapping distributions. This phenomenon is primarily attributed to the fact that the text representations are generated from derivation processes, where inherent noise and semantic shifts lead to a lack of clear category boundaries, thereby failing to provide sufficient discriminative information. After introducing the cross-modal synergistic adapter (bottom row), the text branch features form more compact clusters with distinct boundaries, which in turn promotes the joint improvement of both branches. This confirms the effectiveness of visual injection.

\begin{figure}[t]
  \centering
   \includegraphics[width=0.95\linewidth]{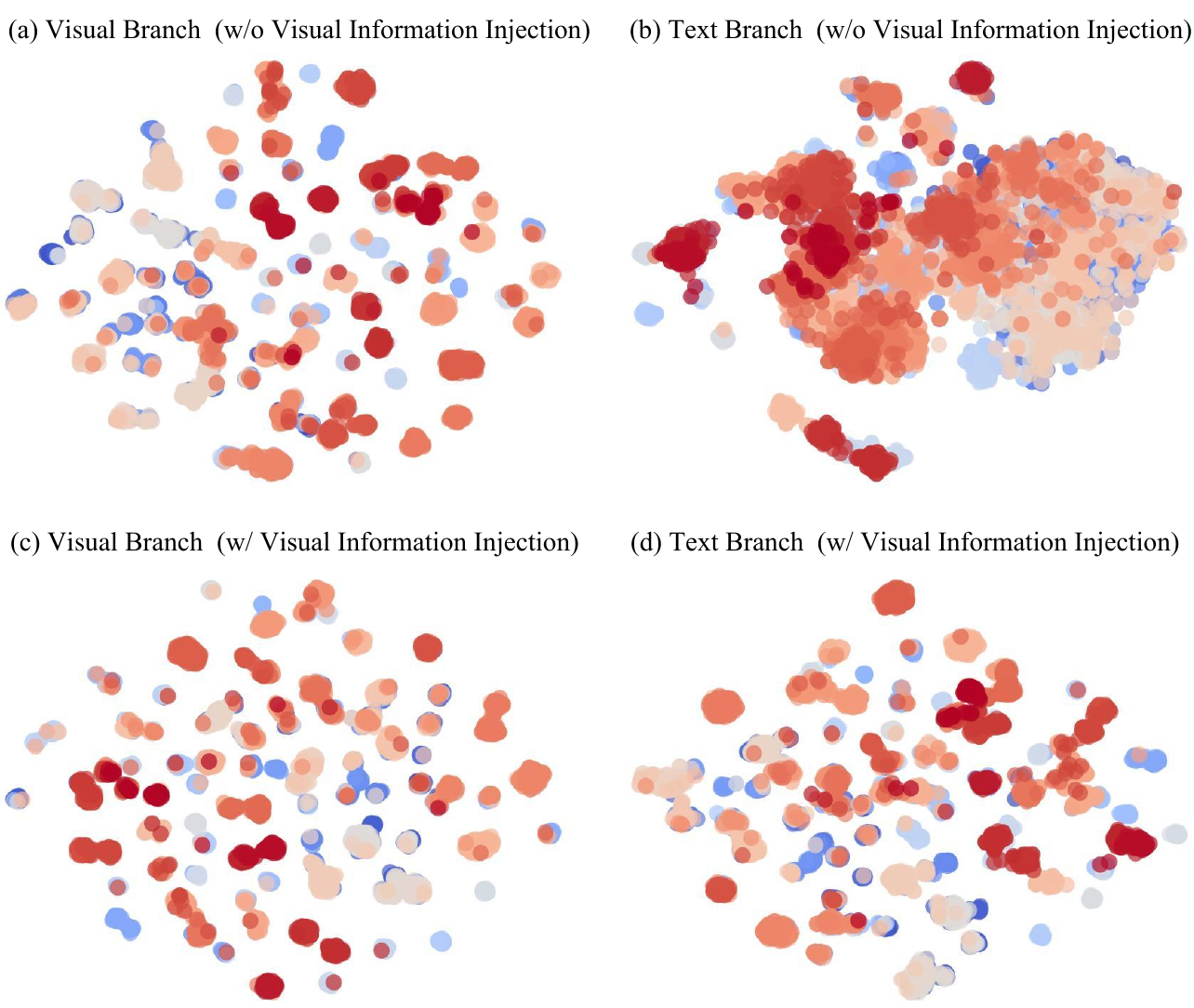}
   \vspace{-2mm}
   \caption{t-SNE visualization of visual and text branch features on Aircraft: (a)(b) without visual injection; (c)(d) with visual injection. The samples on old classes are denoted in blue and the samples on new classes in red.}
   \label{fig:tsne}
\end{figure}
 
\emph{6) Scalability with Unknown Number of Categories.}
In practical open-world scenarios, the true number of categories is often unknown. A common strategy involves using off-the-shelf methods to estimate the category count for initializing the parametric classifiers. Following this paradigm, the category estimation strategy from GET is adopted, yielding 212 classes for CUB-200 and 220 for Stanford Cars \cite{DBLP:conf/cvpr/WangPXYLC25}. Table~\ref{tab:results_est} compares the performance under both known and estimated category count settings. Although both GET and SDBA suffer performance degradation under estimation bias, SDBA consistently maintains its performance advantage over GET across both datasets, demonstrating its robustness in realistic scenarios.
 
\begin{table}[!htbp]
\renewcommand{\arraystretch}{1.3}
\setlength{\tabcolsep}{5pt}
\centering
\caption{Performance comparison on CUB-200 and Stanford Cars under known and estimated category counts ($C$).}
\label{tab:results_est}
\scalebox{1}{
\begin{tabular}{lcccccccc}
\toprule
\multirow{2}{*}{\textbf{Method}} & \multirow{2}{*}{\textbf{Known $C$}} & \multicolumn{3}{c}{\textbf{CUB-200}} & \multicolumn{3}{c}{\textbf{Stanford Cars}} \\
\cmidrule(lr){3-5} \cmidrule(lr){6-8}
& & \textbf{All} & \textbf{Old} & \textbf{New} & \textbf{All} & \textbf{Old} & \textbf{New} \\
\midrule
GET  & \Checkmark       & 77.0 & 78.1 & 76.4 & 78.5 & 86.8 & 74.5 \\
\rowcolor{blue!10}SDBA & \Checkmark & \textbf{79.4} & \textbf{81.6} & \textbf{78.3} & \textbf{80.5} & \textbf{87.8} & \textbf{77.0} \\
\midrule
GET  & \XSolidBrush~(w/ Est.) & 75.6 & 75.9 & \textbf{75.5} & 76.8 & 87.6 & 71.6 \\
\rowcolor{blue!10}SDBA & \XSolidBrush~(w/ Est.) & \textbf{76.7} & \textbf{82.7} & 73.6 & \textbf{78.2} & \textbf{85.5} & \textbf{74.6} \\
\bottomrule
\end{tabular}
}
\end{table}

\emph{7) Impact of Visual Injection Source.}
The visual injection mechanism in CMSA uses the \texttt{[CLS]} token of the visual branch as the guidance signal. We compare this design against using no visual injection (None) and using the mean of all patch tokens excluding the \texttt{[CLS]} token (Mean) on Aircraft, with results reported in Table~\ref{tab:injection_source}. The \texttt{[CLS]} token consistently yields the best performance, while the mean of patch tokens aggregates local spatial features that are less semantically compact, resulting in noisier injection and inferior performance.

\begin{table}[!htbp]
\renewcommand{\arraystretch}{1.3}
\setlength{\tabcolsep}{5.5pt}
\caption{Comparison of visual injection sources on Aircraft
. NML is excluded for fair comparison.}
\label{tab:injection_source}
\centering
\scalebox{1}{
\begin{tabular}{clcccccc}
\toprule
\multirow{2}{*}{\textbf{Dataset}} & \multirow{2}{*}{\textbf{Source}} & \multicolumn{3}{c}{\textbf{Visual Branch}} & \multicolumn{3}{c}{\textbf{Text Branch}} \\
\cmidrule(lr){3-5} \cmidrule(lr){6-8}
& & \cellcolor{blue!10}\textbf{All} & \textbf{Old} & \textbf{New} & \cellcolor{blue!10}\textbf{All} & \textbf{Old} & \textbf{New} \\
\midrule
\multirow{3}{*}{Aircraft}
& None & \cellcolor{blue!10}62.4 & 68.1 & 59.5 & \cellcolor{blue!10}49.5 & 54.6 & 46.9 \\
& Mean & \cellcolor{blue!10}63.0 & 71.6 & 58.7 & \cellcolor{blue!10}63.4 & 71.3 & 59.5 \\
& \cellcolor{blue!10}CLS  & \cellcolor{blue!10}\textbf{64.9} & \cellcolor{blue!10}\textbf{71.3} & \cellcolor{blue!10}\textbf{61.7} & \cellcolor{blue!10}\textbf{64.9} & \cellcolor{blue!10}\textbf{71.1} & \cellcolor{blue!10}\textbf{61.8}
\\
\bottomrule
\end{tabular}
}
\end{table}

\section{Conclusion}
\label{section:conclusion}
This paper addresses multi-modal generalized category discovery from the perspective of cross-modal synergy. Existing dual-branch methods process the two branches in isolation and rely on fixed global anchors, which leaves text-derivation noise unaddressed and yields only coarse-grained cross-modal supervision. To address these limitations, we proposed SDBA, which consists of CMSA that injects visual context into the text adapter to continuously enhance text feature learning during encoding, and NML that complements existing cross-modal mutual learning by aligning local sample neighborhoods across branches, providing fine-grained relational supervision. Comprehensive evaluations on six datasets confirm the effectiveness of our method, and the stable gains obtained on both GET and TextGCD substantiate its scalability as a plug-and-play module. Future directions include exploring finer cross-modal interaction at the local region or token level, and incorporating adaptive fusion weights conditioned on sample difficulty or modality confidence.

\section{Acknowledgement}
This work was supported by the National Key R\&D Program of China (Grant 2022YFF0903300) and the National Natural Science Foundation of China (U22B2048, 62476274, 62206293). Yongqiang Tang is the corresponding author.

\bibliographystyle{IEEEtran}
\bibliography{egbib}

\vfill

\end{document}